\documentclass[letterpaper]{article} % DO NOT CHANGE THIS
\usepackage[preprint]{aaai2027}  % DO NOT CHANGE THIS
\usepackage[hyphens]{url}  % DO NOT CHANGE THIS
\usepackage{graphicx} % DO NOT CHANGE THIS
\usepackage{natbib}  % DO NOT CHANGE THIS AND DO NOT ADD ANY OPTIONS TO IT
\usepackage{caption} % DO NOT CHANGE THIS AND DO NOT ADD ANY OPTIONS TO IT
\usepackage{booktabs,multirow}
\usepackage{algorithm}
\usepackage{algorithmic}
\usepackage{booktabs}
\usepackage{amsmath,amssymb}
\usepackage{multirow}
\newcommand{\ind}{\mathbf{1}}
\newcommand{\Rset}{\mathcal{R}}
\newcommand{\evalfn}{\mathcal{E}}

\newcommand{\Heal}{\mathcal{H}}
\newcommand{\Breach}{\mathfrak{B}}
\newcommand{\dpromote}{\delta_{\mathrm{prom}}}
\newcommand{\dregress}{\delta_{\mathrm{reg}}}
\newcommand{\passk}{\mathrm{pass}^{k}}
\newcommand{\passone}{\mathrm{pass@}1}
\newcommand{\topk}{\textsc{top-}k}
\newcommand{\Mset}{\mathcal{M}}
\newcommand{\Gate}{\mathcal{G}}
\newcommand{\Wspace}{\mathcal{W}}
\newcommand{\gwin}{w}
\newcommand{\gtarget}{\theta}
\newcommand{\ggain}{\gamma_{\mathrm{gain}}}
\newcommand{\gdrop}{\delta_{\mathrm{drop}}}
\newcommand{\outcome}{m_{\mathrm{out}}}
\newcommand{\idx}{\iota}

\title{Self-Healing Harness for Runtime Oversight of Agent Self-Modification}
\author{
Sina Tayebati\textsuperscript{\rm 1},
Divake Kumar\textsuperscript{\rm 1},
Nastaran Darabi\textsuperscript{\rm 1},
Ranganath Krishnan\textsuperscript{\rm 2},
Amit Ranjan Trivedi\textsuperscript{\rm 1}
}

\affiliations{
\textsuperscript{\rm 1} University of Illinois Chicago\\
\textsuperscript{\rm 2} AI Labs, Capital One
}

\begin{document}
\maketitle

% =====================================================================================
\begin{abstract}
LLM agents can change their own future behavior, raising a basic control question of which self-generated changes should be allowed to persist. We formulate this as \emph{admission control for self-modification}. The agent may propose changes to its operating instructions, while an external runtime gate controls persistence. We implement this principle as a model-agnostic self-healing harness that runs a \textbf{Detect}, \textbf{Notice}, \textbf{Heal}, \textbf{Validate} loop around an otherwise unmodified agent. The agent authors candidate behavioral rules in an external workspace, where they receive \emph{provisional} execution authority during evaluation and acquire \emph{persistent} cross-episode authority only after measured improvement on the triggering failure without regression beyond a fixed margin on protected cases. Replay provides matched evidence when available, forward trials provide a weaker fallback, and a corpus-level guard re-tests the accumulated active rule set. Across 16 matched Baseline and Harness runs spanning AppWorld, Terminal-Bench, and $\tau^2$-bench, the gate rejected 383 replay-decided proposals. Of these, 211 ($55\%$) improved their triggering failure while degrading a case that previously worked. This shows that locally beneficial self-modifications can introduce collateral regressions often enough to materially affect gate decisions, providing direct empirical motivation for external admission control. Task-completion score is higher under the Harness in all 16 pairs, with two paired bootstrap intervals excluding zero, while repeated-trial reliability is higher in 12 pairs, tied in 4, and lower in none. Because adaptation modifies the policy-inducing context while leaving model weights fixed, admitted changes remain inspectable, reversible, and compatible with closed-weight models.
\end{abstract}

\section{Introduction}
\label{sec:intro}

Autonomous LLM agents increasingly operate over long horizons, invoke external tools, modify environment state, and retain information across episodes. Persistent deployment therefore lets an agent alter its own future behavior through stored memories, revised operating instructions, or reusable procedures. This introduces the question of which self-generated behavioral updates should be allowed to persist. Improvement on the failure that motivated an update is insufficient by itself, since the same update may degrade behavior that previously succeeded.

We address this by separating update generation from update retention. The agent observes failures and proposes behavioral rules intended to correct them, while an external runtime decides whether each rule becomes persistent, retaining a candidate only when measured execution evidence shows improvement on the motivating failure without regression beyond a fixed margin on protected behavior. This places the persistence decision outside the adapting agent while preserving the agent's ability to generate its own corrections. The mechanism provides external oversight of which self-generated modifications are retained; we do not aim for broad value alignment.

The need for that oversight is visible directly in the agent's own proposals. Across our experiments $383$ candidate rules were rejected by replay-based validation, and $211$ of them ($55\%$) improved the failure that motivated the rule while degrading a protected case that had previously succeeded. A retention policy keyed only to improvement of the triggering failure would have accepted every one of them.

We implement this as \emph{evidence-gated self-modification} inside a self-healing harness (the Harness), a model-agnostic runtime layer around an otherwise unmodified agent that executes a \textbf{Detect}, \textbf{Notice}, \textbf{Heal}, \textbf{Validate} loop: degradation is detected from per-turn score trajectories, the evidence is recorded in an external workspace, the agent inspects it and authors a candidate rule, and validation decides retention by replaying captured failure and protected cases, or by using subsequent executions as weaker forward evidence when replay is unavailable. A corpus-level guard additionally re-tests the accumulated active set, since interactions among individually accepted updates can produce regressions no candidate-level test can see. Because adaptation modifies the policy-inducing context rather than model parameters, retained rules stay explicit, inspectable, reversible, and compatible with closed-weight models.

Deciding \emph{when} to intervene is part of the problem, since an isolated low intermediate score may reflect incomplete progress rather than failure on a long-horizon task. The Harness therefore reads the temporal evolution of per-trace scores rather than individual values, with a per-turn barrier supplying the repeated observations that requires. We measure the operational effect through task completion and execution reliability, distinguishing first-trial success ($\passone$) from all-trials success ($\passk$, the fraction of tasks completed on all $k$ trials), because stochastic agents may succeed on one execution and fail on another. The alignment-relevant claim is narrower. A self-generated update does not become persistent merely because the agent produced it. Our contributions are:
(i) we formulate persistent agent adaptation as \textbf{admission control for self-modification}, where a self-authored update is retained only after demonstrating improvement on its triggering failure while satisfying a multi-metric non-regression constraint on protected behavior; (ii) we develop a model- and framework-independent \textbf{runtime architecture} integrating trajectory-based detection, agent-authored repair, an external rule workspace, replay or forward validation, and corpus-level regression testing, without modifying weights; (iii) we provide \textbf{empirical evidence of collateral regression}, since $211$ of $383$ replay-based rejections improved their triggering failure while degrading protected behavior, so local improvement does not imply reliable persistent adoption; and (iv) we report a \textbf{paired evaluation} over three benchmark families and four models covering task completion, repeated-execution reliability, rule validation, accumulated-rule effects, evaluator robustness, and overhead, including failure modes that persistent adaptation itself introduces.

\section{Background and Related Work}
\label{sec:related}

\noindent\textbf{Self-correction, memory, and verification.}
Reflection and self-refinement let agents revise behavior after failure \citep{madaan2023selfrefine,shinn2023reflexion,gou2024critic}, and memory and experiential-learning systems carry adaptation across episodes through long-term memory \citep{packer2023memgpt,park2023generative} or reusable skills and lessons \citep{wang2023voyager,zhao2024expel}. These determine what experience to generate, retain, or retrieve. Verification methods assess individual outputs \citep{wang2023selfconsistency,lightman2024verify,zheng2023judging}, while guardrails and principle-based steering constrain execution under externally specified policies \citep{tayebati2025cap, tayebati2025learning,rebedea2023nemo, darabi2026eigenshield}. We instead ask whether an agent-authored behavioral update should persist, and condition retention on measured evidence, so the object verified is a modification to the operating context rather than the current answer.

\vspace{2pt}
\noindent\textbf{Persistent adaptation and oversight.} Cross-episode adaptation can accumulate individually useful updates that interact and degrade previously successful behavior, analogous to catastrophic interference in continual learning \citep{mccloskey1989catastrophic}; adapting explicit context rather than parameters keeps each update inspectable and reversible, and a corpus-level guard addresses failures emerging from the accumulated set. Test-time adaptation typically updates parameters from deployment data \citep{sun2020testtime}, whereas we hold $\pi$ fixed and adapt an external rule set, preserving compatibility with API-served models. The mechanism relates to scalable oversight and corrigibility \citep{amodei2016concrete,bowman2022measuring,tayebati2026tracer, kumar2025learnable}, though the gate enforces non-regression only over observed metrics, which here measure task reliability rather than value alignment. Because completion varies across stochastic executions we distinguish $\passone$ from $\passk$, evaluating across digital-work, terminal, and tool-calling environments \citep{trivedi2024appworld,merrill2026terminal,barres2025tau,yao2024taubench}.

\vspace{2pt}
\noindent\textbf{Self-modifying agents and novelty boundary.}
AutoManual builds and revises an environment rule book through interaction \citep{chen2024automanual}, Agent Workflow Memory induces reusable workflows from past trajectories \citep{wang2024awm}, and A-MEM maintains an agent-curated memory store \citep{xu2025amem}. Closer still, STOP retains scaffold modifications on measured meta-utility \citep{zelikman2024stop} and the Darwin G\"{o}del Machine maintains benchmark-evaluated self-modified coding agents \citep{zhang2025dgm}, which establishes that empirical validation of self-modification is not by itself new. What the Harness combines is that the agent authors a \emph{persistent behavioral update} while an external runtime controls retention; admission uses replay or subsequent execution and requires target improvement \emph{together with} protected-case non-regression; updates follow a candidate, active, retired lifecycle with the accumulated set re-tested; and weights stay fixed. Relative to STOP and the Darwin G\"{o}del Machine the distinctions are the protected-case constraint and runtime triggering inside an ongoing deployment rather than an offline improvement search. Table~\ref{tab:gap} compares eleven representative systems on these properties.

\section{The Self-Healing Harness}
\label{sec:harness}

\begin{figure}[t]
\centering
\includegraphics[width=0.97\columnwidth]{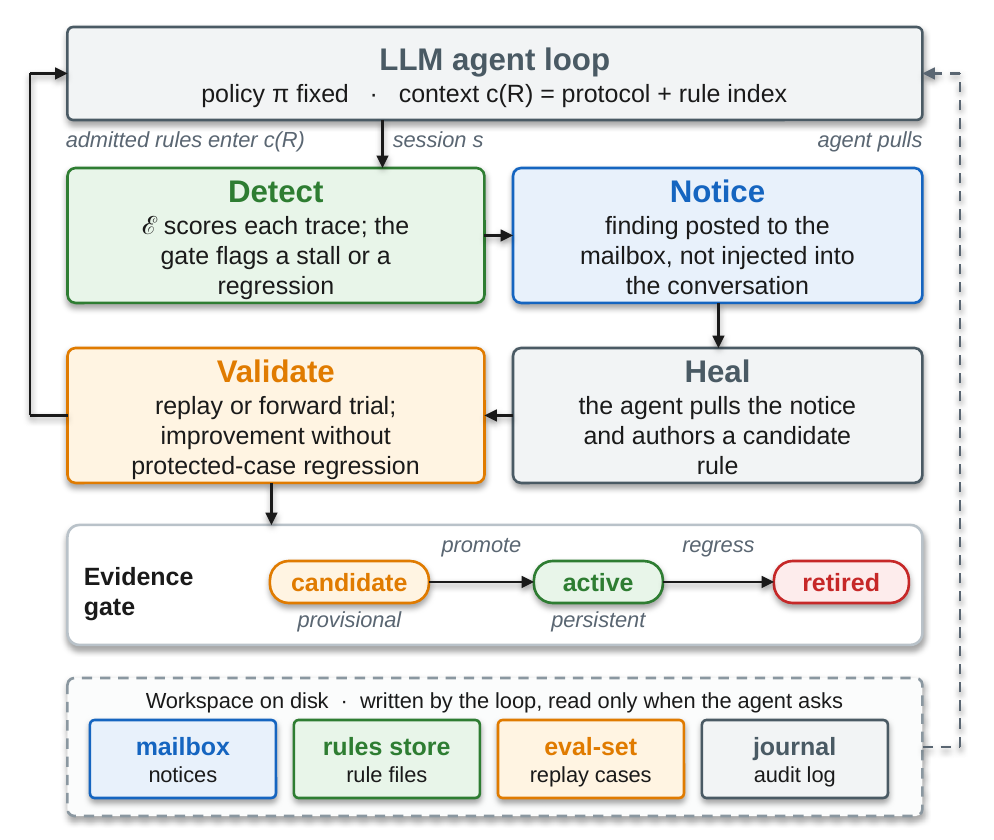}
\caption{\textbf{Harness control boundary.} After each turn, trace-level evaluation detects a stall or regression and escalates when needed (\textbf{Detect}); the finding is written to the external workspace (\textbf{Notice}); the agent retrieves it and authors a candidate rule (\textbf{Heal}); and the candidate remains provisional until replay or forward-trial evidence supports promotion without regression (\textbf{Validate}). A per-turn barrier closes the loop before the next turn, while the pull model keeps workspace contents outside the conversation unless explicitly retrieved.}\vspace{-10pt}
\label{fig:overview}
\end{figure}

\vspace{3pt}
\noindent\textbf{Overview and Control Model.}
The Harness is a runtime envelope around an unmodified agent that lets the agent revise its own operating instructions while enforcing an evidence gate, so only validated revisions become durable: $\pi$ stays constant while the rule set $\Rset$ evolves, making adaptation model-agnostic, auditable, and reversible. Four stages define the control boundary (Figure~\ref{fig:overview}) and give this section its structure. Detect, Notice, and Heal sit on the proposal side of that boundary; only Validate grants persistence, and a per-turn \emph{barrier} closes the loop inside the session in progress.

The mechanism's guarantees are interpretable only against what it governs. We assume an agent that can author natural-language rules, read and write its workspace, retrieve rules into context, and act on any rule in force, but that cannot edit the evaluator, the gate, the replay corpus, or the promotion record, all outside its tool surface. The runtime controls whether a candidate becomes persistent and whether the accumulated active set continues to satisfy non-regression on previously evaluated cases. Each promotion records the case, metric, and improvement that justified it, so every persistent change carries an auditable reason. The agent is assumed non-adversarial: we do not defend against one that circumvents the workspace, persists behavior through another channel, manipulates the evaluator, or authors a change whose harm is invisible to the metric vector, and regressions on behavior the corpus never exercised are undetectable.

\vspace{3pt}
\noindent\textbf{Detect: A Trace-Level Trigger.}
An agent policy $\pi$ acts over a session $s=(x_1,\dots,x_T)$, where each $x_i$ is a per-turn trace. We detect degradation from score trajectories rather than session-level aggregates, since a low intermediate score in a long-horizon task can reflect incomplete progress rather than failure. An evaluation service $\evalfn$ scores each trace on $[0,1]$. Metrics are tiered by cost: inexpensive progress metrics $\Mset_1$ (task completion, coherence) run continuously, while step-level $\Mset_2$ (tool-call and argument correctness), explanatory $\Mset_3$, and outcome $\Mset_0$ are evaluated only after a trajectory trigger (the Appendix).

\noindent\emph{Trajectory gate.}
Tier-1 metrics are judged by the shape of their per-trace series rather than by an absolute score. For metric $j$ the gate $\Gate$ keeps a running peak $\mathrm{pk}_j$ and a counter $z_j$ of traces since the last meaningful gain; a gain occurs when the score exceeds the peak by at least $\ggain$, and otherwise the counter increases and the gate evaluates
\begin{align}
\mathrm{stall}_j(x_i) &= \ind\!\left[\, z_j \ge \gwin \ \wedge\ \mathrm{pk}_j < \gtarget \,\right],
  \label{eq:stall}\\
\mathrm{regr}_j(x_i)  &= \ind\!\left[\, m_j(x_i) < \mathrm{pk}_j - \gdrop \,\right].
  \label{eq:regr}
\end{align}
A gain resets $z_j$ and updates the running peak. Consequently, improving trajectories do not trigger merely because they begin at a low score, plateaus above $\gtarget$ are permitted, and regressions are measured against the best observed state. After a trigger the gate opens a new window. We use $\gwin{=}10$, $\gtarget{=}0.5$, $\ggain{=}0.02$, and $\gdrop{=}0.15$ (Algorithm~\ref{alg:gate}, Table~\ref{tab:hyper}).

\noindent\emph{Escalation and evidence.}
Tier 1 runs on every new trace, and if nothing stalls or regresses evaluation stops there. Otherwise tier 2 scores the latest trace and asks whether the trajectory signal is corroborated by a local execution failure, meaning a step-level or outcome metric below its absolute threshold. Only a corroborated finding, a \texttt{breach}, is recorded as a replayable case for validation; an uncorroborated one is surfaced to the agent as a \texttt{trend} but never becomes promotable evidence. Corroboration is therefore required before a signal becomes evidence for a persistent change, and RQ5 reports its effect when the evaluator itself is biased.

\noindent\emph{Outcome evidence.}
\label{sec:verifier}
Tier-1 and tier-2 metrics are judged proxies, so when a deployment provides a direct success signal the Harness admits it as an optional \emph{outcome verifier} that returns a score or abstains, and validation prefers this evidence over any judged proxy. Abstention is explicit, since a verifier that scores only completed tasks cannot report mid-session.

\vspace{3pt}
\noindent\textbf{Per-Turn Barrier.}
After each turn, execution waits until new traces are evaluated and any resulting notice is posted. This allows a repair authored at turn $i$ to affect turn $i{+}1$, and supplies the repeated within-session observations the trajectory gate requires. Synchronizing only once per task would give a single observation per metric series, leaving the stall detector inoperative. Per-turn synchronization is therefore a precondition of the trigger rather than an optimization.

\vspace{3pt}
\noindent\textbf{The External Rule Workspace.}
Learned rules must stay available without forcing the whole corpus into context every turn, since full inlining raises both token cost and competition for attention. The Harness therefore separates the \emph{index} of learned rules from their \emph{content}:
\begin{equation}
  c(\Rset) = c_0 \oplus \idx(\Rset),
  \label{eq:context}
\end{equation}
where $c_0$ is a fixed root document holding the standing protocol and tool interface and $\idx(\Rset)$ is a compact directory of rule files, their sizes, and pending notices. No rule bodies appear in this context: rule text enters context only when the agent reads a scope or searches the workspace, so the standing context grows with the \emph{number} of rule files rather than the volume of what has been learned, though retrieved text does occupy the conversation once pulled, which is the mechanism behind the token growth in RQ4. This is a \emph{pull model}: notices, rules, traces, replay cases, and the audit journal live in a workspace $\Wspace$, and none of it is injected into the conversation. The agent reaches them through explicit tool calls, so workspace contents remain external until retrieved. 

\vspace{3pt}
\noindent\textbf{Heal: Agent-Authored Behavioral Updates.}
Repair is authored by the agent and retention is controlled by the runtime. After retrieving a notice and inspecting the flagged trace, the agent writes a candidate behavioral rule $r$, which becomes persistent only if it passes the evidence test below. The agent therefore determines what to propose while the runtime determines what persists. The Appendix states the update operator and rule metadata; Algorithm~\ref{alg:heal} gives the per-turn procedure.

\vspace{3pt}
\noindent\textbf{Validate: Evidence-Gated Self-Modification.}
The gate controls cross-episode persistence. A candidate remains provisional and may influence behavior while its effect is measured, since a candidate under trial must be in force for the trial to measure anything, but persists across episodes only after promotion. Promotion records the supporting evidence. Validation uses replay when suitable prior cases are available and a weaker forward trial otherwise (Algorithm~\ref{alg:validate}, Appendix).

\noindent\emph{Replay scoring and case selection.}
Replay re-scores the captured session on $\Mset_1\cup\Mset_2$, the same metrics that can produce a triggering notice, so the gate tests a candidate against the signal that motivated the repair rather than a broad aggregate; $\Mset_3$ is excluded as explanatory. The verdict is read against the most authoritative target available per case, a direct verifier outcome if there is one, else the metric the rule declares, else the metric named by the triggering signature, so abstention never blocks validation. Each round replays at most three \emph{failure} cases, matched to the candidate's signature $\mathrm{sig}(r)$, and at most two \emph{protected} cases.

A protected case is a captured session paired with the metrics on which it scored above threshold. Because capture is triggered at the session level while scores are recorded per metric, a session that entered the corpus on one failing metric still contributes its passing metrics as protected behavior during replay. The non-regression condition therefore rejects a candidate that repairs its triggering metric while degrading another that previously succeeded, which makes the test a cross-case non-regression test rather than a target-only improvement test.
At most two protected cases are replayed per round and we do not rank them for semantic proximity to the candidate, which is the conservative reading of the mechanism: protection is necessarily incomplete and side effects outside the sampled cases can escape detection. Protected-set construction is therefore a first-order determinant of what the gate can see, and \emph{Limitations} lists relevance-ranked selection as an important untested variant.

\noindent\emph{The admission test.}
Each selected session is replayed under the candidate context $c(\Rset\cup\{r\},\cdot)$ and re-scored. Writing $\Delta_j(s_i)$ for the replay-minus-original difference on metric $j$ of case $s_i$, a candidate must improve its target metric $\hat{\jmath}$ on at least one failure case,
\(
  \exists i:\Delta_{\hat{\jmath}}(s_i)\geq\dpromote,
\)
while satisfying non-regression across all replayed cases and all measured metrics,
\(
  \forall i,\forall j:\Delta_j(s_i)>-\dregress,
\)
with $\dpromote=\dregress=0.05$. These two predicates are the admission rule. The admission criterion gives precedence to non-regression: any non-regression violation therefore causes rejection irrespective of target improvement. Inconclusive replay falls back to the forward trial after at most three attempts. The test is a bounded empirical procedure, not a certificate of causation, since agents are stochastic and at most five sessions are replayed, so a delta near the margin can reflect sampling variation and repeated-replay agreement of verdicts is unmeasured.

\noindent\emph{Forward trial, and why it is weaker.}
When replay is unavailable the gate instead measures whether the candidate reduces future occurrences of its triggering failure, comparing the flagged rate over the $n$ sessions following activation against the pre-activation rate $p_0$ and promoting only if the failure stops or its rate falls by at least $\dpromote$ (the Appendix). The study uses $n=3$ so a trial completes inside one run, with $p_0=1$ when no history exists. This path is materially weaker: three sessions are unmatched evidence, there is no protected-case component so collateral damage cannot be detected at all, and the default $p_0=1$ makes promotion comparatively easy. RQ3 reports how the two paths were distributed across runs.

\vspace{3pt}
\noindent\textbf{Regression Guard for Behavioral Drift.}
Candidate-level validation tests individual updates against sampled evidence, while regressions can emerge as the active set accumulates. The guard therefore replays the whole evaluation corpus under the current active rules, re-scores each case on $\Mset_1\cup\Mset_2$, and fails the corpus if any previously evaluated case drops by at least $\dregress$ on any measured metric, improvement elsewhere not offsetting a regression. Within the coverage of that corpus the guard bounds degradation of previously evaluated behavior at $\dregress$ per metric. Three limits bound the guarantee. It covers only properties in the metric vector, which here measure task performance, so it establishes non-regression in measured reliability rather than value alignment, although safety or permission signals routed through the verifier would fall under the same guard. Replay detects only regressions the corpus exercises. And it inherits the reliability of $\evalfn$, including the evaluator noise of RQ5.

% =====================================================================================
\section{Experimental Setup}
\label{sec:setup}

\noindent\textbf{Protocol.}
We use a paired A/B protocol in which \textbf{Baseline}, the unmodified agent loop, and \textbf{Harness}, the same loop with Detect, Notice, Heal, and Validate enabled, share the agent implementation, model, prompts, task set, and execution order and differ only in self-healing. \textbf{PandaProbe} \citep{Tayebati_pandaprobe_2026} supplies the tracing and evaluation infrastructure with identical metric definitions, gate parameters, and thresholds in both arms. Benchmark-native graders determine $\passone$ and $\passk$ and supply the continuous signal behind $\bar{s}$, while trace-level evaluations drive Harness adaptation, keeping the evaluator separate from every reported outcome measure.

\vspace{3pt}
\noindent\textbf{Benchmarks and models.}
We evaluate \textbf{AppWorld} \citep{trivedi2024appworld} for long-horizon application and API tasks, \textbf{Terminal-Bench} \citep{merrill2026terminal} for command-line tasks, and \textbf{$\tau^2$-bench} \citep{barres2025tau} for tool-using dialogue. The four splits are AppWorld \texttt{test\_normal} ($168$ tasks), Terminal-Bench \texttt{sample@2.0} ($10$), and $\tau^2$ \texttt{airline} ($50$) and \texttt{retail} ($114$), each evaluated with \textsc{gpt-5.6-terra}, \textsc{gpt-5.6-luna}, \textsc{claude-haiku-4.5}, and \textsc{claude-sonnet-5} using $k{=}4$ trials, a $100$-turn budget per task, and seed $1$, yielding $16$ matched pairs (Table~\ref{tab:config}, Appendix). Trial-to-trial stochasticity is retained because the evaluated endpoints do not expose temperature control.

\vspace{3pt}
\noindent\textbf{Online adaptation.}
The Harness remains active throughout each run, allowing earlier tasks to affect later behavior through proposed, validated, promoted, and retired rules, so matched pairs use the same execution order and all outstanding validation jobs are settled before archival. Each split is processed once as a continuous adaptation stream without a separate learning and evaluation partition.

\vspace{3pt}
\noindent\textbf{Metrics and statistics.}
Each task uses the benchmark's native binary success criterion, with $\passone$ denoting first-trial success and $\passk$ success on all four trials. For $\bar{s}$, we compute paired per-task Harness-minus-Baseline differences and report their mean with a $95\%$ percentile-bootstrap interval over $10{,}000$ task-level resamples and a two-sided bootstrap $p$-value, resampling tasks rather than trials because repeated trials of one task are dependent \citep{efron1979bootstrap}. Intervals containing zero are treated as directional, and formal metric and interval definitions appear in the Appendix.

% =====================================================================================
\section{Results}
\label{sec:results}
\noindent\textbf{Scope and Organization.} We report 16 matched Baseline--Harness pairs across three benchmarks, four splits, and four models, with $k{=}4$ trials per task and seed $1$. Each comparison uses only task-trials completed by both arms under the same benchmark, dataset, model, seed, and $k$ configuration. We organize the results around five questions. \textbf{RQ1}: does gated persistent adaptation improve reliability? \textbf{RQ2}: do self-generated updates require admission control? \textbf{RQ3}: does the gate discriminate among proposals, and how does the validation path affect outcomes? \textbf{RQ4}: does accumulated adaptation introduce additional risk? \textbf{RQ5}: what are the robustness and overhead of continuous runtime oversight?

\vspace{3pt}
\noindent\textbf{RQ1: Does gated adaptation improve reliability?}
We report a \emph{task-completion score} $\bar{s}$ as the primary continuous metric: each benchmark's own continuous signal placed on one $[0,1]$ scale, namely AppWorld's and Terminal-Bench's passing-test fraction and $\tau^2$'s per-component reward, averaged over all trials. It is the highest-resolution quantity these benchmarks expose. $\passone$ and $\passk$ are reported alongside it. Table~\ref{tab:headline} gives the primary comparison. The clearest single result is AppWorld with \textsc{gpt-terra}, the study's largest paired evaluation at $166$ tasks, where the Harness raises $\bar{s}$ from $0.701$ to $0.723$ ($\Delta=+0.022$) with a bootstrap $95\%$ CI of $[+0.006,+0.040]$ that excludes zero ($p=0.010$), and raises $\passone$ from $0.018$ to $0.054$. The second interval that excludes zero is $\tau^2$ retail with \textsc{claude-sonnet-5}, $\Delta\bar{s}=+0.032$, CI $[+0.003,+0.061]$, $p=0.03$.

Across all $16$ pairs the Harness improves $\bar{s}$, leads on $\passk$ in $12$ and ties in $4$, and leads on $\passone$ in $14$ with one tie and one Baseline lead. Only two $\bar{s}$ intervals exclude zero, so the remaining effects are directional. The $\tau^2$ family shows the smallest gains, between $+0.012$ and $+0.032$, and RQ3 identifies a validator limitation specific to part of that family.

\begin{table*}[t]
\centering
\footnotesize
\setlength{\tabcolsep}{7pt}
\renewcommand{\arraystretch}{1.08}

\caption{Headline results for matched Baseline (B) and Harness (H) runs.
$\bar{s}$ denotes task-completion score. CI$_{95}$ and $p$ are the paired
task-level bootstrap interval and two-sided $p$-value for H$-$B in $\bar{s}$.
Bold H entries indicate numerical Harness leads, while bold CI$_{95}$/$p$
entries indicate intervals excluding zero. All remaining differences are directional.
Rates are computed over the task-trials present in both arms, so the denominator behind
$\passone$ and $\passk$ can differ within a row.}\vspace{-10pt}
\label{tab:headline}

\begin{tabular}{@{}llrccclcccc@{}}
\toprule
&&&
\multicolumn{3}{c}{Task-completion score $\bar{s}$} &
\multicolumn{2}{c}{$\passone$} &
\multicolumn{2}{c}{$\passk$} \\
\cmidrule(lr){4-6}
\cmidrule(lr){7-8}
\cmidrule(lr){9-10}
Benchmark & Model & $n$
& B & H & CI$_{95}$ / $p$
& B & H
& B & H \\
\midrule

\multirow{4}{*}{\textbf{AppWorld}}
& gpt-terra
& 166 & 0.701 & \textbf{0.723}
& \textbf{[+0.006,+0.040] / .01}
& 0.018 & \textbf{0.054}
& 0.000 & 0.000 \\

& gpt-luna
& 168 & 0.658 & \textbf{0.662}
& [-0.01,+0.02] / .54
& 0.000 & \textbf{0.006}
& 0.000 & 0.000 \\

& claude-haiku-4.5
& 168 & 0.566 & \textbf{0.590}
& [-0.06,+0.10] / .58
& 0.000 & \textbf{0.143}
& 0.000 & \textbf{0.143} \\

& claude-sonnet-5
& 166 & 0.714 & \textbf{0.727}
& [-0.005,+0.032] / .17
& 0.030 & \textbf{0.048}
& 0.006 & \textbf{0.012} \\

\addlinespace[3pt]
\cmidrule(lr){1-10}
\addlinespace[2pt]

\multirow{4}{*}{\textbf{Terminal-Bench}}
& gpt-terra
& 10 & 0.450 & \textbf{0.525}
& [-0.13,+0.33] / .65
& 0.444 & \textbf{0.556}
& 0.333 & \textbf{0.444} \\

& gpt-luna
& 10 & 0.275 & \textbf{0.300}
& [-0.10,+0.15] / .72
& 0.300 & \textbf{0.400}
& 0.100 & \textbf{0.200} \\

& claude-haiku-4.5
& 10 & 0.229 & \textbf{0.375}
& [-0.28,+0.52] / .42
& 0.600 & 0.600
& 0.000 & \textbf{0.200} \\

& claude-sonnet-5
& 10 & 0.398 & \textbf{0.417}
& [-0.41,+0.43] / .90
& 0.571 & 0.429
& 0.000 & \textbf{0.333} \\

\addlinespace[3pt]
\cmidrule(lr){1-10}
\addlinespace[2pt]

\multirow{8}{*}{\textbf{$\tau^2$-bench}}
& gpt-terra (retail)
& 114 & 0.681 & \textbf{0.704}
& [-0.004,+0.050] / .09
& 0.384 & \textbf{0.411}
& 0.125 & \textbf{0.143} \\

& gpt-terra (airline)
& 50 & 0.724 & \textbf{0.741}
& [-0.019,+0.054] / .34
& 0.500 & \textbf{0.540}
& 0.360 & \textbf{0.400} \\

& gpt-luna (retail)
& 114 & 0.650 & \textbf{0.662}
& [-0.02,+0.04] / .47
& 0.339 & \textbf{0.357}
& 0.099 & \textbf{0.117} \\

& gpt-luna (airline)
& 50 & 0.690 & \textbf{0.704}
& [-0.02,+0.05] / .42
& 0.460 & \textbf{0.480}
& 0.280 & \textbf{0.320} \\

& claude-haiku-4.5 (retail)
& 114 & 0.632 & \textbf{0.651}
& [-0.009,+0.047] / .18
& 0.304 & \textbf{0.321}
& 0.071 & \textbf{0.080} \\

& claude-haiku-4.5 (airline)
& 50 & 0.682 & \textbf{0.696}
& [-0.02,+0.05] / .44
& 0.440 & \textbf{0.460}
& 0.320 & \textbf{0.340} \\

& claude-sonnet-5 (retail)
& 114 & 0.669 & \textbf{0.701}
& \textbf{[+0.003,+0.061] / .03}
& 0.366 & \textbf{0.384}
& 0.107 & 0.107 \\

& claude-sonnet-5 (airline)
& 50 & 0.710 & \textbf{0.740}
& [-0.006,+0.066] / .10
& 0.440 & \textbf{0.480}
& 0.320 & 0.320 \\

\bottomrule\vspace{-10pt}
\end{tabular}
\end{table*}

\begin{figure}[t]
\centering
\includegraphics[width=\columnwidth]{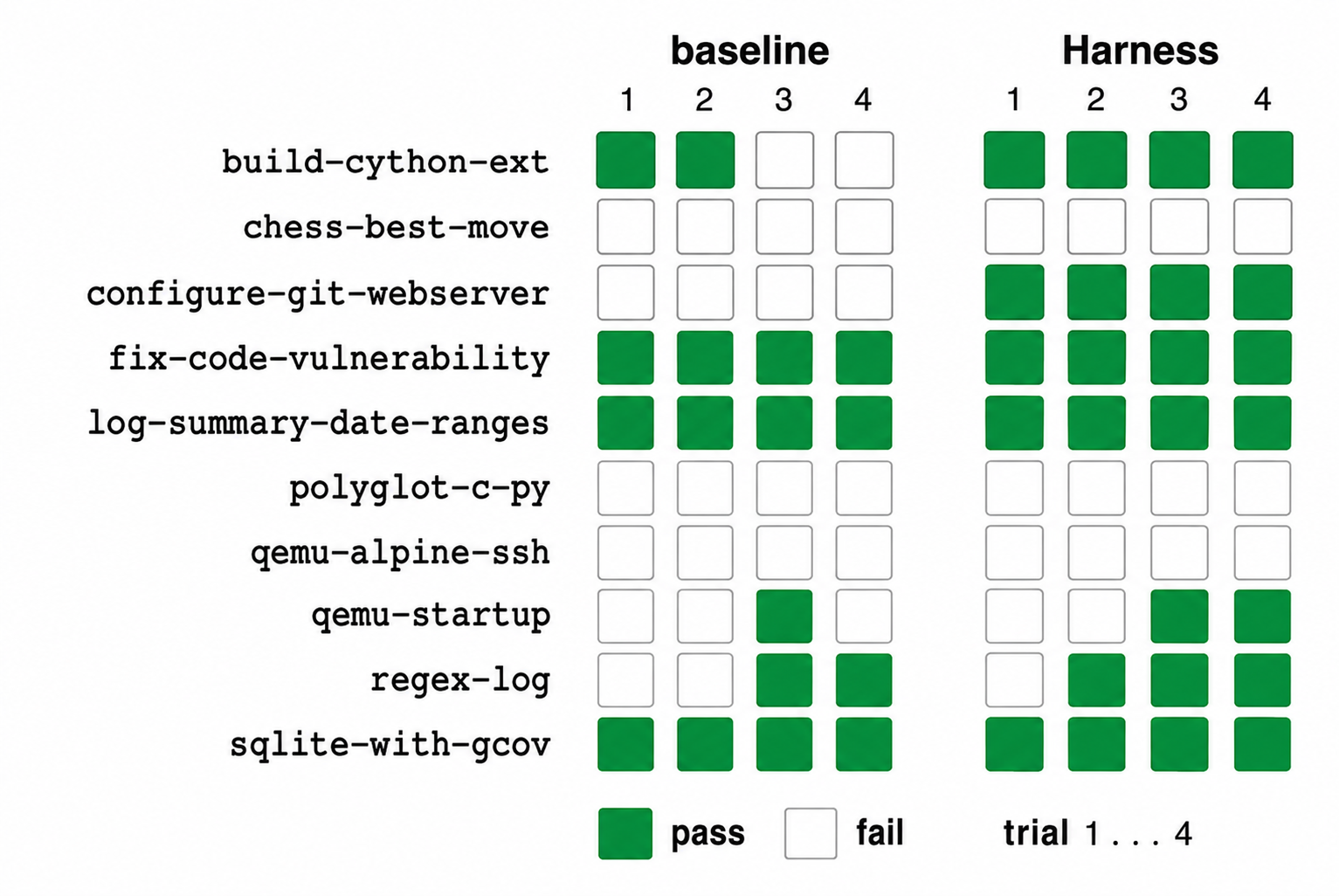}
\caption{\textbf{Per-task trial outcomes} on Terminal-Bench with \textsc{terra}; filled cells denote successful trials. Under the Harness, \texttt{build-cython-ext} and \texttt{regex-log} become consistently successful after intermittent Baseline performance, while \texttt{configure-git-webserver} improves from zero to four successes. Overall, four tasks succeed on all four trials under the Harness versus three under Baseline.}\vspace{-10pt}
\label{fig:trials}
\end{figure}

\noindent\emph{Repeatability.}
Figure~\ref{fig:trials} illustrates the per-task pattern on Terminal-Bench, where several intermittent Baseline successes become repeatable under the Harness. Because Terminal-Bench contains only $10$ tasks, its $\passk$ values move in coarse steps. The gains do not generally concentrate in $\passk$ relative to $\passone$: only $3$ of $16$ pairs satisfy $\Delta\passk>\Delta\passone$ (Table~\ref{tab:signature}, Appendix), so we claim consistent improvement in repeated-trial success rather than a repeatability-specific effect.

\vspace{3pt}
\noindent\textbf{RQ2: Do self-generated updates need admission control?}
Across the $16$ Harness runs the agent proposed $2{,}306$ rules, of which $1{,}244$ were promoted and $1{,}062$ retired (Table~\ref{tab:lifecycle}, Appendix), so roughly $46\%$ of proposals were rejected. The more informative statistic is \emph{why}.

\noindent\emph{Conflict prevention.}
Among $383$ replay-decided retirements, $211$ ($55\%$) improved their triggering failure but regressed protected behavior (Table~\ref{tab:gate}). Observed regressions include a $0.60$ decrease in $\mathit{task\_completion}$ and a $0.25$ decrease in $\mathit{outcome\_correct}$. Replay-decided verdicts are a subset of the totals above, the remainder having been resolved by forward trial, and Terminal-Bench contributes none because replay is unavailable there (RQ3).
These proposals were rejected because the non-regression condition overrode measured local improvement, so the majority of rejected self-modifications here were \emph{locally correct and harmful to protected behavior}. Under the observed replay evidence, a retention rule based only on triggering-case improvement would have admitted all $211$. This measurement does not depend on the end-task comparison, since it characterizes the agent's proposals under a fixed evidence test rather than the reliability difference in RQ1. The $211$ cases are also detected conflicts under sparse protection, not an estimate of total collateral-regression incidence. Each validation round examines at most two unranked protected cases, so harmful effects outside that small sample remain unobserved. The result therefore establishes that collateral regression occurs frequently enough to be detected under limited coverage, while providing no guarantee that promoted rules are free of unseen regressions.

% requires: booktabs, multirow

\begin{table}[t]
\centering
\footnotesize
\setlength{\tabcolsep}{3.0pt}
\renewcommand{\arraystretch}{1.08}

\caption{Replay-based gate decisions, pooled over the runs in which replay resolved the verdict; Terminal-Bench contributes none, since replay is unavailable there. Promotions are grouped by the metric that improved beyond the admission margin. Among retirements, 55\% improved the triggering failure but regressed protected behavior.}\vspace{-10pt}
\label{tab:gate}

\begin{tabular}{@{}lrrl@{}}
\toprule
Decision basis & Count & Share & Verdict \\
\midrule

\texttt{task\_completion}
    & 200 & 48\% & \multirow{5}{*}{\textbf{Promote} (420)} \\
\texttt{tool\_correctness}
    &  73 & 17\% & \\
\texttt{outcome\_correct}
    &  73 & 17\% & \\
\texttt{coherence}
    &  58 & 14\% & \\
\texttt{argument\_correctness}
    &  15 &  4\% & \\

\addlinespace[3pt]
\midrule

Protected-case regression
    & \textbf{211} & \textbf{55\%}
    & \multirow{2}{*}{\textbf{Retire} (383)} \\
No triggering-case improvement
    & 172 & 45\% & \\

\bottomrule\vspace{-15pt}
\end{tabular}
\end{table}

\vspace{3pt}
\noindent\textbf{RQ3: Does the gate discriminate, and does the evidence path matter?}
\noindent\emph{Selectivity.}
Promotion rates vary substantially across benchmarks and validation paths, from a minority on AppWorld to $0.41$--$0.63$ on $\tau^2$, with the per-run rates in Table~\ref{tab:lifecycle}. Replay-decided promotions span all five evaluated metrics rather than concentrating on a single signal (Table~\ref{tab:gate}).

\noindent\emph{Natural variation in validation path.}
Replay availability differs structurally across benchmarks (Table~\ref{tab:validator}), so these differences are observational rather than an ablation. Terminal-Bench cannot replay because validating one candidate would require rebuilding and rerunning the task environment, and its four runs correspondingly show zero retirements. The $\tau^2$ airline runs instead suffered a replay-budget unit mismatch, the budget specified in agent turns while $\tau^2$ consumes it as message hops, which truncated many replays before the relevant behavior was exercised.

Both statistically resolved improvements occur among runs with high replay share. Replay availability is confounded by benchmark and model, however, so this association does not establish a performance advantage for replay. It does establish that part of the $\tau^2$ corpus received weaker validation than intended and that Terminal-Bench received no replay-based regression testing at all. Replay share therefore also indicates how much of a run's validation activity received the protected-case non-regression test, which the forward path lacks.

\begin{table}[t]
\centering
\footnotesize
\setlength{\tabcolsep}{2.4pt}
\renewcommand{\arraystretch}{1.06}
\caption{Validation path by run. Replay \% is the share of validation activity resolved by replay rather than forward trial, and R/F reports the corresponding replay and forward event counts. These are per validation event and do not correspond one-to-one with the rule counts of Table~\ref{tab:lifecycle}. Terminal-Bench does not support replay and therefore has structural $0\%$ rows. Variation in replay availability is observational rather than a controlled ablation. Bold indicates replay shares above $80\%$.}

\label{tab:validator}

\begin{tabular}{@{}llrrrr@{}}
\toprule
Benchmark & Model & Replay \% & R/F & $\Delta\bar{s}$ & $\Delta\passk$ \\
\midrule

\multirow{4}{*}{\textbf{AppWorld}}
& terra      & $\mathbf{87\%}$ & 104/15 & +0.022 & +0.000 \\
& luna       & $\mathbf{90\%}$ & 189/21 & +0.004 & +0.000 \\
& haiku-4.5  & $\mathbf{84\%}$ & 168/32 & +0.024 & +0.143 \\
& sonnet-5   & $\mathbf{86\%}$ &  93/15 & +0.013 & +0.006 \\

\addlinespace[2pt]

\multirow{4}{*}{\textbf{Terminal-Bench}}
& terra      & 0\% & 0/206 & +0.075 & +0.111 \\
& luna       & 0\% & 0/324 & +0.025 & +0.100 \\
% PLACEHOLDER-TBHAIKU: provisional value for Terminal-Bench / claude-haiku-4.5 pass^k.
% The recorded 1.000 exceeded pass@1 = 0.600, which the metric definition forbids. 0.200
% (1 of 5) is consistent with pass@1 = 3/5 and with the other Terminal-Bench rows.
% REPLACE with the recomputed value; three sites are marked with this tag.
& haiku-4.5  & 0\% & 0/280 & +0.146 & +0.200 \\
& sonnet-5   & 0\% & 0/230 & +0.019 & +0.333 \\

\addlinespace[2pt]

\multirow{4}{*}{\textbf{$\tau^2$ airline}}
& terra      & 72\% & 118/46 & +0.017 & +0.040 \\
& luna       & 69\% & 151/68 & +0.014 & +0.040 \\
& haiku-4.5  & 64\% & 132/74 & +0.014 & +0.020 \\
& sonnet-5   & 58\% &  59/43 & +0.030 & +0.000 \\

\addlinespace[2pt]

\multirow{4}{*}{\textbf{$\tau^2$ retail}}
& terra      & $\mathbf{89\%}$ & 246/30 & +0.023 & +0.018 \\
& luna       & $\mathbf{92\%}$ & 357/29 & +0.012 & +0.018 \\
& haiku-4.5  & $\mathbf{88\%}$ & 332/45 & +0.019 & +0.009 \\
& sonnet-5   & $\mathbf{91\%}$ & 211/21 & +0.032 & +0.000 \\

\bottomrule
\end{tabular}
\end{table}

\vspace{3pt}
\noindent\textbf{RQ4: Does accumulated adaptation create new risk?}
Candidate-level admission tests one update against the available evidence; it does not establish whether the accumulated adaptation state remains reliable, and our data show the second question is not implied by the first. Table~\ref{tab:learning} tracks per-task $\Delta\bar{s}$ across execution quartiles alongside active-set size, each Harness task differenced against its Baseline counterpart so difficulty is controlled and execution order is not read as learning.
It covers the configurations for which quartile-resolved telemetry is available. Those five rows carry three positive and two negative slopes, so the subset is not assembled around the negative result reported below. For most configurations the advantage is present early rather than emerging after accumulation: on AppWorld with \textsc{gpt-terra} it stays positive in all four quartiles, and both $\tau^2$ datasets show positive headline effects and positive slopes as the corpus grows. AppWorld with \textsc{gpt-luna} is the informative exception, declining monotonically from $+0.017$ in Q1 to $-0.016$ in Q4 ($p=0.022$) while the active set grows from $13$ to $70$ rules and input tokens per turn rise from $1.50\times$ to $2.52\times$ Baseline. Every rule passed the gate individually, so the harm is a property of the corpus rather than of any member.

Two mechanisms are consistent with this pattern: context dilution, as an expanding corpus competes for attention, and rule interference, when individually valid rules give conflicting guidance; the present data do not distinguish them.

\noindent\emph{Why the corpus guard did not prevent this.}
The guard exists for exactly this class of effect, but three scope limits mean it could not have registered this particular decline.
\emph{Coverage}: the guard replays only captured \texttt{breach} cases, whereas the decline is measured over the full task stream. \emph{Granularity}: the guard detects per-case drops of at least $\dregress=0.05$, while the observed Q4 deficit is a smaller aggregate shift distributed across tasks. \emph{Mechanism}: if degradation arises from retrieval dilution as the corpus grows, replaying a stored case may not reproduce the live retrieval state that caused it. The guard therefore bounds regression on replayed cases but does not control aggregate drift over uncaptured behavior. Addressing this failure mode requires controls over the accumulated adaptation state in addition to per-candidate admission. Candidate admission and corpus-level oversight therefore address different failure scales, the first at the level of individual updates and the second at the level of their accumulated interactions.

% requires: booktabs, multirow

\begin{table}[t]
\centering
\footnotesize
\setlength{\tabcolsep}{1pt}
\renewcommand{\arraystretch}{1.06}

\caption{Learning dynamics for the five configurations with quartile-resolved telemetry.
Q1--Q4 report paired $\Delta\bar{s}$ over execution quartiles; Rules gives the mean active-rule
count in Q1 and Q4. Slope is per task in units of $10^{-3}$, with permutation $p$-values from
$5{,}000$ shuffles.}
\label{tab:learning}

\begin{tabular}{@{}llrrrrrr@{}}
\toprule
Benchmark & Model
& Q1 & Q2 & Q3 & Q4
& Rules & slope ($p$) \\
\midrule

\multirow{2}{*}{AppWorld}
& terra
& +.031 & +.001 & +.040 & +.018
& 10$\to$47 & +.03 (.86) \\

& luna
& +.017 & +.009 & +.006 & -.016
& 13$\to$70 & -.27 (.02) \\

\addlinespace[2pt]

Term-Bench
& terra
& +.250 & +.500 & -.125 & -.125
& 7$\to$7 & -74.2 (.09) \\

\addlinespace[2pt]

$\tau^2$ airline
& luna
& -.010 & -.010 & +.063 & +.012
& 11$\to$81 & +1.23 (.26) \\

$\tau^2$ retail
& luna
& -.003 & +.029 & -.025 & +.047
& 36$\to$195 & +.20 (.67) \\

\bottomrule\vspace{-20pt}
\end{tabular}
\end{table}

\vspace{3pt}
\noindent\textbf{RQ5: How robust and how expensive is oversight?}
\vspace{3pt}
\noindent\emph{Mechanism verification.}
Median metric-series length ranges from $6$ to $11$ across runs, confirming repeated within-session observations, while tier-2 escalation rates range from $0.29$ to $0.83$. These are operational checks rather than ablations; full telemetry appears in Table~\ref{tab:telemetry} in the Appendix.

\vspace{3pt}
\noindent\emph{Evaluator noise.}
During AppWorld development, trace judges incorrectly penalized read-only Harness meta-tools, with $14\%$ of sampled rationales blaming meta-tool calls and $31$ spurious zero scores. Neutrality guidance reduced these to $0.17\%$ and $1$ over $4{,}800$ sampled rationales (Table~\ref{tab:noise}, Appendix). Both arms of every reported pair are scored by the same judge prompts, so an evaluator revision cannot differentially favor either arm of a comparison.
Because detection and admission are separate, such trigger noise does not directly grant persistence.

\vspace{3pt}
\noindent\emph{Cost and overhead.}
Mean context usage is $1.25\times$ Baseline and agent turns increase by $0.0$ to $1.8$ per trial. Task-execution spend rises from $\$177.06$ to $\$211.46$, with $\$92.90$ additional repair cost, yielding $1.72\times$ Baseline overall. The dominant overhead is latency, with wall-clock time increasing by roughly $1.7$ to $7\times$, primarily from synchronous trace evaluation, which is a material deployment constraint for latency-sensitive applications. Per-run accounting appears in Tables~\ref{tab:cost} and~\ref{tab:costbenefit} in the Appendix.

\vspace{3pt}
\noindent\textbf{Limitations.}
The primary missing comparison is between evidence-gated and immediate admission of the same agent-authored rules (arm B$'$), with the downstream corpus guard fixed. The Baseline--Harness comparison therefore does not isolate the effect of pre-admission validation. This limits claims about the benefit of gating, but not the RQ2 observation that protected-case regressions occur directly among the agent's proposals under the fixed replay test. The no-barrier, inlined-rule, tier-3, replay-only, and forward-trial-only arms were also not run, and the protocol measures online adaptation rather than held-out transfer. Only two of 16 paired intervals exclude zero, the remaining effects are directional, and all runs use seed 1 although adaptation is path dependent. The gate is a bounded empirical test over at most five replayed sessions, with repeated-replay stability and sensitivity to $\dpromote$, $\dregress$, and $\gwin$ unmeasured. Forward trials use only three unmatched sessions and no protected-case test, leaving Terminal-Bench and part of $\tau^2$ airline supported by weaker evidence. Non-regression is further limited by protected-set coverage, since at most two unranked protected cases are replayed per round, so the 211 conflicts in RQ2 are detected conflicts rather than an estimate of total incidence. The guarantees cover only measured properties, here task reliability, and cannot exclude unmeasured safety violations or failures from irreversible actions. Synchronous oversight adds roughly $1.7$--$7\times$ wall-clock latency, while RQ4 shows that per-candidate admission does not control degradation of the accumulated rule set. Evaluation is limited to three benchmark families, text-based rules, and a non-adversarial agent within its tool surface, leaving richer persistent modifications and adversarial circumvention open.

\section{Conclusion}
\label{sec:conclusion}

We formulate persistent agent self-modification as admission control, where the agent proposes behavioral updates and an external runtime controls persistence through replay-based improvement, multi-metric non-regression, a weaker forward-trial fallback, and corpus-level re-testing. The central empirical result is that 211 of 383 replay-decided rejections ($55\%$) improved their triggering failure while degrading protected behavior; under the observed replay evidence, a target-only retention rule would have admitted all 211. Because each validation round inspects at most two protected cases, these are detected conflicts under sparse coverage rather than an estimate of total incidence. We also observe declining performance as individually validated rules accumulate, indicating that per-candidate admission alone does not control corpus-level degradation. These results motivate external evidence-gated control over self-modification.

\section*{Disclosure}

Capital One does not use nor has plans to use this methodology in any capacity
(e.g., audits, credit decision explainability, anti-discrimination testing).

\bibliography{references}

% =====================================================================================
% TECHNICAL APPENDIX
% =====================================================================================
% \clearpage
\appendix
\section{Technical Appendix}
\label{sec:appendix}

\noindent This appendix carries material deferred from the main text. It is organized so that
each part answers a question a reader of the main text may raise, in the order those questions
arise. \emph{Experiments not run} states the arms we designed but did not execute, so the gap
between the thesis and the comparison is explicit rather than implied. \emph{Metric and
interval definitions} gives the formal definitions behind the reliability metrics and the
bootstrap interval. \emph{Mechanism details} specifies, item by item, the implementation of
every mechanism whose scientific content is stated in the Harness section. \emph{Algorithms}
gives the per-turn loop, the evidence gate, and the trajectory-gate update in full.
\emph{Notation} and \emph{Hyperparameters} are the complete symbol reference and the complete
constant table, the latter listing every package default beside the value actually used.
\emph{Comparison to prior mechanisms} positions eleven representative systems against the five
properties of controlled self-improvement, and \emph{Additional result tables} holds the
per-run tables behind aggregate numbers reported in the results. Nothing here changes a claim
made in the main text.

\subsection{Experiments not run}
\label{sec:app-notrun}
Arms that were designed but not executed, expanding the main text discussion of what the study
does not isolate.

\vspace{2pt}
\noindent\textbf{Three configurations.}
It is worth separating three settings that are easy to conflate:
\begin{itemize}\itemsep1pt
\item \textbf{B}: candidate validation before admission, plus the corpus-level guard. This is what we ran.
\item \textbf{B$'$}: immediate admission of authored rules, with the same corpus-level guard downstream.
\item \textbf{B$''$}: immediate admission with no corpus guard, that is, genuinely uncontrolled persistence.
\end{itemize}
B$'$ is the comparison that directly isolates the effect of pre-admission validation while holding the downstream corpus guard fixed. We are explicit that B$'$ is \emph{not} ungated adaptation: an external evidence-based regression check still operates after admission, so the contrast is between validating before admission and validating only afterwards. B$''$ is the fully uncontrolled setting, and we name it for completeness rather than proposing it as the primary comparison, since removing both layers would confound the two mechanisms.

\vspace{2pt}
\noindent\textbf{The B$'$ arm.}
Configuration B$'$ is the Harness with candidate validation disabled, so that an authored rule enters the active set immediately. Everything else is held identical to B: the same detection, notices, agent, repair prompts, rule authoring, retrieval, corpus guard, task order, and benchmark. The measurements that would separate B from B$'$ are final task completion, $\passone$ and $\passk$, the number of self-authored rules retained, the number of retained rules that later regress a protected case under the same corpus guard, the magnitude of those regressions, and performance as a function of execution order. The prediction implied by RQ2 needs care, because adaptation is path dependent. Under the observed B trajectories, automatic admission would have retained the $211$ proposals that B rejected for protected-case regression. An actual B$'$ run would not encounter those same proposals: once an early rejected rule is retained, subsequent failures, notices, and authored rules diverge. The testable claim is therefore distributional rather than case-matched. Immediate admission should accumulate rules faster and should show a higher rate and larger magnitude of downstream protected-case regression under the identical guard, together with worse late-run paired performance if those regressions are consequential. That formulation also anticipates the obvious objections to reading RQ2 as sufficient on its own: the rejected rules might prove harmless in live execution, some protected-case drops might be evaluator noise, faster accumulation might outweigh the regressions, and the corpus guard might catch the damage later anyway. The last of these is precisely why B$'$ retains the guard, since a difference that survives an identical downstream check is attributable to the admission test. A controlled B versus B$'$ comparison is therefore the most direct experiment for isolating the effect of the candidate-level admission test.

\vspace{2pt}
\noindent\textbf{Other arms not run.}
The no-barrier, inlined-rule, tier-3-enabled, replay-only, and forward-trial-only arms were also not run. The barrier evidence in RQ5 is mechanism verification, and the replay-share variation in RQ3 is observational and confounded. Neither substitutes for the corresponding controlled arm.

\vspace{2pt}
\noindent\textbf{Held-out transfer.}
The protocol measures online reliability during adaptation, in one continuous pass over each split. It does not measure whether the accumulated rule corpus transfers. The corresponding experiment is to run the Harness over an adaptation stream, freeze the final validated corpus, and evaluate Baseline against frozen-Harness on held-out tasks, which would separate reusable behavioral improvement from useful intervention within the same stream.

\subsection{Metric and interval definitions}
\label{sec:app-defs}
Formal statements of the reliability metrics and the confidence interval used throughout the
results, deferred from the experimental setup. For task set $T$ with binary benchmark outcomes
$X_{t,i}\in\{0,1\}$ over $k$ trials,
\begin{equation}
\textstyle
\passone = \frac{1}{|T|}\sum_{t\in T}X_{t,1},\;\;
\passk = \frac{1}{|T|}\sum_{t\in T}\prod_{i=1}^{k}X_{t,i},
\label{eq:passdef}
\end{equation}
so $\passone$ is first-trial success while $\passk$ requires success on every trial. For the
paired per-task difference $d_t$ in task-completion score, with $R=10{,}000$ bootstrap
resamples over tasks,
\begin{equation}
\textstyle
  \mathrm{CI}_{95} =
  \bigl[\,Q_{2.5}(\{\bar{d}^{(\ast r)}\}_{r=1}^{R}),\;
  Q_{97.5}(\{\bar{d}^{(\ast r)}\}_{r=1}^{R})\,\bigr].
  \label{eq:bootstrap}
\end{equation}

\subsection{Mechanism details}
\label{sec:app-mech}
Implementation-level specifications for the mechanisms whose scientific content is stated in
the main text. Nothing here changes a claim; each item is the formal or operational form of a
sentence in the main text.

\vspace{3pt}
\noindent\textbf{Metric ladder}, from \emph{Detect}.
$\Mset_1$ (\emph{progress}, always on) is task completion and coherence, evaluated on every
new trace; coherence is embedding-based, which makes this the cheapest tier. $\Mset_2$
(\emph{step-level diagnosis}) is tool-call correctness and argument correctness, evaluated
only on the latest trace once $\Mset_1$ flags the trajectory. $\Mset_3$ (\emph{explanatory},
opt-in) is planning adherence, plan quality, and step efficiency, used only to enrich a
confirmed finding and never as a trigger. $\Mset_0$ (\emph{outcome}) is the score from the
optional developer-defined verifier. The ladder is also the cost-control mechanism: tier 1
scores all of a turn's new traces in one batched request, tier 2 scores only the latest trace
after a gate fire, and tier 3 is off by default. Because model-judged metrics consume the
full trace, confining expensive diagnosis to flagged trajectories is what makes continuous
monitoring tractable.

\vspace{3pt}
\noindent\textbf{Absolute breaches and severity}, from \emph{Detect}.
For metrics with an explicit threshold $\tau_j$,
\begin{equation}
  b_j(x) = \ind\!\left[\,m_j(x) \ne \varnothing \ \wedge\ m_j(x) < \tau_j\,\right],
  \label{eq:breach}
\end{equation}
with pending scores excluded. Absolute thresholds apply only to outcome and step-level
metrics,
\begin{equation}
  \mathrm{breach}_j(x) = b_j(x)\ \wedge\ \mathrm{tier}(j)\in\{0,2\},
  \label{eq:absbreach}
\end{equation}
so tier-1 metrics are governed only by the stall and regression conditions and tier-3
metrics never produce a breach. The severity assigned to a fired trajectory condition is then
\begin{equation}
\mathrm{sev} =
\begin{cases}
\texttt{breach} & \exists\, j\in\Mset_2 \cup \Mset_0:\ \mathrm{breach}_j(x_T),\\
\texttt{trend}  & \text{otherwise,}
\end{cases}
\label{eq:sev}
\end{equation}
evaluated on the latest trace $x_T$. Each fired condition is emitted as a signature of the
form \texttt{condition:metric}, such as \texttt{stall:task\_completion} or
\texttt{breach:tool\_correctness}.

\vspace{3pt}
\noindent\textbf{Barrier timeout}, from the per-turn barrier.
The barrier has a dedicated timeout budget. If it expires, evaluation continues
asynchronously and the turn is marked timed out, so a slow evaluator adds latency without
blocking task execution or changing the correctness of already-completed work.

\vspace{3pt}
\noindent\textbf{Outcome verifier}, from \emph{Detect}.
Given the session state, the verifier returns either a score or an abstention,
\begin{equation}
  \outcome(s) \in [0,1] \cup \{\varnothing\},
  \label{eq:verifier}
\end{equation}
so a verifier that can score only completed tasks returns $\varnothing$ on intermediate turns.
Valid outcomes enter the ladder as tier-0 metrics against a strict threshold
$\tau_{\mathrm{out}}=0.9$. Verifier failures are isolated from the agent loop: exceptions are
caught and the turn proceeds without an outcome score. Because verification depends only on
the turn's end state, it runs concurrently with trace-level metric evaluation.

\vspace{3pt}
\noindent\textbf{Scopes}, from the rule workspace.
Rules are partitioned into free-form \emph{scopes}, each backed by one rule file. The Harness
provides a \texttt{global} scope for broad trajectory-level lessons and a \texttt{scoped}
default for step-level findings, and the agent may create more specific topic scopes. Notices
suggest an initial scope, which the agent may override when authoring or reorganizing rules.
New scopes appear in $\idx(\Rset)$ automatically. The ranking function applied at retrieval
time is the relevance score there below.

\vspace{3pt}
\noindent\textbf{Heal as an update operator}, from \emph{Heal}.
At step $t$, given the breaches $\Breach_t$ identified by $\evalfn$,
\begin{equation}
  \Rset_{t+1} = \Heal(\Rset_t,\Breach_t)
  = \mathrm{gate}\bigl(\Rset_t,\mathrm{propose}(\Rset_t,\Breach_t)\bigr),
  \label{eq:heal}
\end{equation}
where \emph{propose} is carried out by the agent and \emph{gate} is the evidence test of
the evidence gate. Candidates are de-duplicated against the live rule set using normalized
rule text.

\vspace{3pt}
\noindent\textbf{Target metric selection}, from \emph{Validate}.
The verdict is read against
\begin{equation}
  \hat{\jmath} \;=\; \operatorname{first-available}\bigl(\,\outcome,\;
  \mathrm{met}(r),\; \mathrm{met}(\mathrm{sig}(r))\,\bigr),
  \label{eq:target}
\end{equation}
where $\mathrm{met}(\cdot)$ denotes the metric named by a rule or by its triggering signature.
Availability is resolved per case, so verifier abstention does not block validation.

\vspace{3pt}
\noindent\textbf{Replay delta}, from \emph{Validate}.
For replayed case $s_i$ and metric $j$,
\begin{equation}
  \Delta_j(s_i)=m'_j(s_i)-m_j(s_i),
  \label{eq:delta}
\end{equation}
with $m'$ the replay score and $m$ the score recorded when the case was captured.

\vspace{3pt}
\noindent\textbf{Forward-trial predicate}, from \emph{Validate}.
Let $p_0$ be the pre-activation rate of the rule's metric-specific signature family
(\texttt{stall}, \texttt{regression}, or \texttt{breach}), estimated as flagged sessions
divided by total sessions, with $p_0=1$ when no prior window exists. After $n$ subsequent
sessions under the candidate, the empirical flagged rate $\hat p$ must satisfy
\begin{equation}
  \hat{p}=0 \quad\text{or}\quad \hat{p}\leq p_0-\dpromote.
  \label{eq:forward}
\end{equation}
Otherwise the rule is retired, and until enough sessions accumulate, validation stays pending.

\vspace{3pt}
\noindent\textbf{Eval-case capture}, from \emph{Validate}.
Each captured case stores the replay input together with its baseline trace scores. Advisory
\texttt{trend} notices are not captured, which keeps the replay corpus focused on diagnosed
failures and keeps the deltas in the replay delta interpretable against a recorded
baseline.

\vspace{3pt}
\noindent\textbf{Worked example}, from \emph{Validate}.
Suppose an agent repeatedly applies a destructive action to the wrong file and proposes the
rule ``verify the target path before any destructive action.'' If replay improves the target
metric by $0.30$ without violating the non-regression condition, the rule is promoted. If the same
rule repairs the original failure but degrades a protected case by more than $\dregress$, it
is retired. The second case is not hypothetical; it is the modal rejection in our runs
(RQ2).

\vspace{3pt}
\noindent\textbf{Corpus regression classification}, from the regression guard.
The guard classifies each replayed case from its per-metric deltas,
\begin{equation}
  \mathrm{cls}(s_i) =
  \begin{cases}
    \text{regressed} & \exists\, j:\ \Delta_j(s_i) \le -\dregress,\\[2pt]
    \text{improved}  & \text{else if } \exists\, j:\ \Delta_j(s_i) \ge \dpromote,\\[2pt]
    \text{unchanged} & \text{otherwise,}
  \end{cases}
  \label{eq:regcls}
\end{equation}
so regression takes precedence over improvement and the rule set passes only if no replayed
case is classified as regressed.

\vspace{3pt}
\noindent\textbf{Healing action space}, from the rule workspace.
Healing is exposed through a small self-diagnostic interface. The agent can list, read, and
acknowledge notices; inspect traces; and read, search, add, retire, or query the status of
rules. Each operation returns a structured response and fails gracefully, so a diagnostic
error does not interrupt task execution. The full set of tunable constants appears in
Table~\ref{tab:hyper}.

\subsection{Relevance score for search}
Deferred from the rule workspace. This score orders the agent's \emph{search} results and
narrows the active rules rendered inside one scope file when a query is supplied. It does not
select what enters the prompt by default, since the context decomposition carries no rule bodies.
Each rule $r$ carries tag tokens $T_r$ and text tokens $X_r$ drawn from its rule text,
rationale, and metric. The query token set $Q$ is assembled from pending notice signatures and
metric names plus any caller-supplied terms, case-folded and tokenized on
\texttt{[a-z0-9\_]+} with length $\ge 2$. The relevance of $r$ to $Q$ is
\begin{equation}
  \mathrm{score}(r,Q) = \frac{2\,\lvert Q \cap T_r\rvert + \lvert Q \cap (X_r \setminus T_r)\rvert}{\max(1,\lvert Q\rvert)} ,
  \label{eq:score}
\end{equation}
so a tag-token match is weighted $2.0$ and a non-tag text-token match $1.0$, normalized by
query size. A scope file rendered under a query keeps its $\topk$ active rules by score
($k=8$), with ties broken by score, then recency, then id, and appends a one-line note naming
the search operation so that nothing becomes unreachable:
\begin{equation}
  \rho_{\sigma}(\Rset,Q) = \mathrm{render}\!\Bigl( \operatorname*{top-}k_{\;r\in \Rset_\sigma}\ \mathrm{score}(r,Q) \Bigr),
  \label{eq:rho}
\end{equation}
where $\Rset_\sigma$ is the active subset of scope $\sigma$. \emph{Candidate} rules are exempt
from this cutoff and always render in full, which is the retrieval-level expression of the
provisional execution authority described under \emph{Validate}: a candidate under trial has
to be in force for the trial to measure anything, so narrowing it away would starve its own
validation.

\subsection{Algorithms}
\noindent The three procedures the main text refers to, in full.

\vspace{3pt}
\noindent\textbf{Per-turn self-heal loop.}
Algorithm~\ref{alg:heal} states the full per-turn loop deferred from \emph{Heal}; the
evidence gate it invokes is Algorithm~\ref{alg:validate}.

\begin{algorithm}[h]
\caption{Self-Heal Loop (one turn)}
\label{alg:heal}
\textbf{Input}: rule set $\Rset$, service $\evalfn$, workspace $\Wspace$, gate $\Gate$
\begin{algorithmic}[1]
\STATE compose context $c(\Rset)$ via the context decomposition \COMMENT{protocol $+$ index; no rule text}
\STATE run one agent turn; new traces $X \gets (x_{i},\dots,x_{T})$ not yet scored
\STATE $\Breach \gets \varnothing$
\STATE \COMMENT{tier 1: every new trace, one batched service call}
\FORALL{$x \in X$ in chronological order}
    \FORALL{$j \in \Mset_1$}
        \STATE fold $m_j(x)$ into $\Gate$ \COMMENT{Algorithm~\ref{alg:gate}}
        \STATE add \textsc{stall}/\textsc{regr} verdicts to $\Breach$
    \ENDFOR
\ENDFOR
\IF{$\Breach \neq \varnothing$}
    \STATE \COMMENT{tier 2: last trace only}
    \STATE add $\{\,\mathrm{breach}_j(x_T) : j \in \Mset_2\,\}$ to $\Breach$
      \COMMENT{the absolute-breach predicate}
    \IF{any tier-2 breach \AND tier 3 enabled}
        \STATE score $\Mset_3$ on $x_T$ \COMMENT{explanatory only}
    \ENDIF
\ENDIF
\STATE $\outcome \gets$ verifier$(s)$ if wired \COMMENT{concurrent; may abstain}
\IF{$\Breach \neq \varnothing$ and not deduplicated/cooled-down}
    \STATE $\mathrm{sev} \gets$ the severity rule; post notice to mailbox $\subset \Wspace$
    \IF{$\mathrm{sev} = \texttt{breach}$ and capture enabled}
        \STATE capture the session as a replayable eval-case
    \ENDIF
\ENDIF
\STATE \textsc{ValidateCandidates}($\Rset$) \COMMENT{Algorithm~\ref{alg:validate}}
\STATE \textbf{barrier}: wait until the above has landed \COMMENT{the per-turn barrier}
\STATE \COMMENT{pull model: the agent may now pull notices on its own initiative}
\IF{agent pulls a notice}
    \STATE agent inspects the flagged trace, reads the suggested scope file
    \STATE agent authors candidate rule $r$; $\Rset \gets \Rset \cup \{r\}$
      \COMMENT{provisional authority only}
\ENDIF
\STATE \textbf{return} $\Rset$
\end{algorithmic}
\end{algorithm}

\vspace{3pt}
\noindent\textbf{The evidence gate.}
Algorithm~\ref{alg:validate} states the gate deferred from \emph{Validate}. It is driven
from the turn hook on every handled report, as a single-flight background round, so it never
blocks a turn and any failure inside it degrades to a log line rather than an exception in the
host loop.

\begin{algorithm}[h]
\caption{\textsc{ValidateCandidates}: the evidence gate}
\label{alg:validate}
\begin{algorithmic}[1]
\FORALL{candidate $r \in \Rset$}
    \IF{replay fn.\ available \AND attempts$(r) < 3$}
        \STATE select $\le 3$ failure cases matching $\mathrm{sig}(r)$, $\le 2$ protected wins
        \STATE replay each under $c(\Rset\cup\{r\})$; re-score on $\Mset_1\cup\Mset_2$
        \STATE deltas $\Delta_j(s_i)$ via the replay delta; target $\hat{\jmath}$ via
               the target-selection rule
        \IF{$\exists\,i,j:\Delta_j(s_i)\le-\dregress$}
            \STATE \textbf{retire} $r$ \COMMENT{the non-regression condition: regression dominates}
        \ELSIF{$\exists\,i:\Delta_{\hat{\jmath}}(s_i)\ge\dpromote$}
            \STATE \textbf{promote} $r$ \COMMENT{the improvement condition}
        \ELSIF{no conclusive failure case}
            \STATE mark \emph{pending}; attempts$(r)\!+\!+$
        \ELSE
            \STATE \textbf{retire} $r$ \COMMENT{no measured improvement}
        \ENDIF
    \ELSE
        \STATE observe $n$ sessions; form $p_0,\hat{p}$ \COMMENT{forward-trial fallback}
        \IF{$\hat{p}=0$ \OR $\hat{p}\le p_0-\dpromote$}
            \STATE \textbf{promote} $r$ \COMMENT{the forward-trial predicate}
        \ELSIF{sessions observed $\ge n$}
            \STATE \textbf{retire} $r$
        \ELSE
            \STATE \emph{pending}
        \ENDIF
    \ENDIF
\ENDFOR
\end{algorithmic}
\end{algorithm}

\vspace{3pt}
\noindent\textbf{The trajectory gate.}
Deferred from \emph{Detect}. Algorithm~\ref{alg:gate} is the complete per-metric update.
It is a pure function of the stored state $(\mathrm{pk}_j,z_j)$ and the new value, which is why
a whole turn's traces can be folded in chronological order in a single write to the history
store. The three properties claimed in the main text are visible here: a gain resets the counter
and raises the peak, so a climbing session never fires; the stall test compares the \emph{peak}
against $\gtarget$ rather than the current value, so a session that reached its target may
plateau; and both conditions reset the counter after firing, so one stall yields one alert.

\begin{algorithm}[h]
\caption{$\Gate$: trajectory-gate update for metric $j$ on one trace}
\label{alg:gate}
\textbf{Input}: state $(\mathrm{pk}_j, z_j)$, value $v = m_j(x_i)$\\
\textbf{Output}: updated state, verdict $\in\{\textsc{ok},\textsc{stall},\textsc{regr}\}$
\begin{algorithmic}[1]
\IF{$\mathrm{pk}_j = \varnothing$ \OR $v \ge \mathrm{pk}_j + \ggain$}
    \STATE $\mathrm{pk}_j \gets \max(\mathrm{pk}_j, v)$;\ \ $z_j \gets 0$
    \STATE \textbf{return} \textsc{ok} \COMMENT{progress: reset the window}
\ENDIF
\STATE $z_j \gets z_j + 1$
\STATE $\mathrm{stall} \gets [\,z_j \ge \gwin \ \wedge\ \mathrm{pk}_j < \gtarget\,]$
  \COMMENT{Eq.~\eqref{eq:stall}}
\STATE $\mathrm{regr} \gets [\,v < \mathrm{pk}_j - \gdrop\,]$ \COMMENT{Eq.~\eqref{eq:regr}}
\IF{$\mathrm{stall}$ \OR $\mathrm{regr}$}
    \STATE $z_j \gets 0$ \COMMENT{fire once, then reopen the window}
\ENDIF
\STATE \textbf{return} verdict from $(\mathrm{stall},\mathrm{regr})$
\end{algorithmic}
\end{algorithm}

State is persisted per (session, metric) alongside the score series, so it survives process
restarts. A session's series is exactly the sequence of tier-1 samples the per-turn barrier
caused to be collected, which is why its length is an operational check (RQ5) rather than an
implementation detail.

\subsection{Full notation}
Table~\ref{tab:notation} is the complete symbol reference. The main text introduces each
symbol where it is first used and does not repeat the table.

\begin{table}[h]
\centering\small
\caption{Complete notation. Where a default and a study value differ, the study value is
given in Table~\ref{tab:hyper}.}
\label{tab:notation}
\begin{tabular}{@{}ll@{}}
\toprule
Symbol & Meaning \\
\midrule
$\pi$ & the (fixed) agent policy wrapped \\
$s=(x_1,..,x_T)$ & a session (one task trial); its traces \\
$x_i$ & the $i$-th trace (one agent turn) \\
$\evalfn$ & trace-level evaluation service \\
$m_j(x)$ & metric $j$ on trace $x$; $\varnothing$ if pending \\
$\Mset_0$ & $\{\outcome\}$: verifier outcome (tier 0) \\
$\Mset_1$ & progress metrics, every trace (tier 1) \\
$\Mset_2$ & step-level metrics, last trace (tier 2) \\
$\Mset_3$ & explanatory metrics, opt-in (tier 3) \\
$\tau_j$ & absolute threshold, metric $j$ (def.\ $0.5$) \\
$\tau_{\mathrm{out}}$ & outcome threshold (def.\ $0.9$) \\
$b_j(x)$ & point-breach indicator on metric $j$ \\
$\Gate$ & trajectory gate \\
$\mathrm{pk}_j, z_j$ & gate state: peak, traces since gain \\
$\gwin,\gtarget$ & gate window, target (def.\ $5$, $0.5$; study $\gwin{=}10$) \\
$\ggain,\gdrop$ & gain / drop margins ($0.02$, $0.15$) \\
$\Rset$ & rule set (active $\cup$ candidate) \\
$r$ & a rule (text, rationale, scope, tags) \\
$c(\Rset)$ & policy-inducing context \\
$\idx$ & rule-\emph{index} operator \\
$\Wspace$ & filesystem workspace \\
$\Heal$ & self-heal operator on the rule set \\
$\Breach_t$ & conditions observed at turn $t$ \\
$\hat{\jmath}$ & promotion target metric \\
$\dpromote,\dregress$ & promote / regress margins ($0.05$) \\
$n$ & forward-trial window (def.\ $5$; study $3$) \\
$p_0,\hat{p}$ & baseline / trial flagged rate \\
$k$ & trials per task in evaluation \\
\bottomrule
\end{tabular}
\end{table}

\subsection{Hyperparameters}
Table~\ref{tab:hyper} lists every tunable constant with both its package default and the value
used in the reported runs. Exactly one constant differs across benchmarks, the replay turn
budget, so the gate constants, the evidence-gate margins, and the trial window are held fixed
for all three benchmarks and all four models. No constant was tuned per benchmark, which is why
\emph{Limitations} reports threshold sensitivity as an open cost rather than a result.

\begin{table}[h]
\centering\footnotesize
\caption{Harness hyperparameters: package default vs.\ the value used in every reported run.
\textbf{Boldface} marks a study override. Eval-case \emph{capture} defaults off and is on in
the Harness arm throughout the run, so replay validation has cases to work with; tier~3 stays
off because each model-judged metric re-reads the whole trace. Rows marked $^{\dagger}$ are
harness-level constants with no package default; $^{\ast}$ marks the only value that differs
across benchmarks. The forward-trial window is lowered to $3$ so that a rule can complete a
trial inside a single run, and the barrier and poll budgets are raised because trace
evaluation is model-judged and takes minutes.}
\label{tab:hyper}
\setlength{\tabcolsep}{3.5pt}%
\resizebox{\columnwidth}{!}{%
\begin{tabular}{@{}llcc@{}}
\toprule
Symbol / knob & Meaning & Default & Study \\
\midrule
\multicolumn{4}{@{}l}{\emph{Trigger and trajectory gate}}\\
$\gwin$ & gate window (traces w/o new high) & $5$ & $\mathbf{10}$ \\
$\gtarget$ & gate target & $0.5$ & $0.5$ \\
$\ggain$ & gain counted as progress & $0.02$ & $0.02$ \\
$\gdrop$ & drop from peak $=$ regression & $0.15$ & $0.15$ \\
$\tau_j$ & absolute threshold (tiers $0,2$) & $0.5$ & $0.5$ \\
$\tau_{\mathrm{out}}$ & outcome-verifier threshold & $0.9$ & $0.9$ \\
-- & tier 3 enabled & off & off \\
-- & traces listed per turn & $50$ & $50$ \\
\midrule
\multicolumn{4}{@{}l}{\emph{Barrier and cost}}\\
-- & barrier budget (s) & $180$ & $\mathbf{1080}$ \\
-- & score poll: interval $\times$ tries & $1{\times}20$ & $\mathbf{5{\times}200}$ \\
-- & global concurrent evaluations & $4$ & $4$ \\
\midrule
\multicolumn{4}{@{}l}{\emph{Rules}}\\
$k$ & \topk\ actives per queried scope & $8$ & $8$ \\
-- & max live rules (active$+$cand.) & \multicolumn{2}{c}{uncapped} \\
\midrule
\multicolumn{4}{@{}l}{\emph{Evidence gate}}\\
$n$ & forward-trial window & $5$ & $\mathbf{3}$ \\
$\dpromote$ & promote margin & $0.05$ & $0.05$ \\
$\dregress$ & regress margin & $0.05$ & $0.05$ \\
-- & max replay attempts & $3$ & $3$ \\
-- & failure / win cases replayed & $3\,/\,2$ & $3\,/\,2$ \\
-- & replay timeout (s) & $300$ & $\mathbf{180}$ \\
-- & replay turns$^{\dagger\ast}$ & -- & $\mathbf{15}$; $\tau^2\,\mathbf{100}$ \\
-- & replay env-wait grace (s) & $0$ & $\mathbf{900}$ \\
-- & validation round budget (s) & $0$ & $\mathbf{600}$ \\
-- & regression sample & \multicolumn{2}{c}{$0=$ whole set} \\
-- & validation / capture & on / off & on / \textbf{on} \\
\midrule
\multicolumn{4}{@{}l}{\emph{End-of-run settlement}}\\
-- & settle timeout / poll$^{\dagger}$ & -- & $\mathbf{1080\,/\,10}$ \\
\midrule
\multicolumn{4}{@{}l}{\emph{Run design}}\\
-- & trials per task $k$ & \multicolumn{2}{c}{$4$} \\
-- & seed & \multicolumn{2}{c}{$1$ of $\{1,2,3\}$} \\
-- & per-task agent turns & \multicolumn{2}{c}{$100$} \\
\bottomrule
\end{tabular}}
\end{table}

\subsection{Comparison to prior mechanisms}
Table~\ref{tab:gap} expands the novelty boundary stated in related work. The columns are five
properties of controlled self-improvement. \emph{P}, persistent, meaning the change survives
across episodes. \emph{S}, self-generated, meaning authored by the agent rather than a human.
\emph{V}, validated by measurement before the change is retained. \emph{R}, regression-guarded,
meaning retention is conditioned on not degrading behavior that previously succeeded.
\emph{T}, triggered by a measured runtime signal during deployment rather than by a fixed
schedule, an offline search loop, or an author's decision about when to intervene.

The rows are chosen to make the claim harder rather than easier. Reflexion, ExpeL, AutoManual,
Agent Workflow Memory, and A-MEM all let an agent write durable operating knowledge, and none
conditions retention on a measured non-regression test. STOP and the Darwin G\"{o}del Machine
are the closest comparators, because both score self-modifications empirically before keeping
them, so \emph{V} is satisfied. They differ on the remaining two columns: neither conditions
retention on protected cases that previously passed, and both run as offline improvement
searches rather than as a runtime loop inside an ongoing deployment. The Harness combination is
\emph{V} and \emph{R} and \emph{T} together on an agent-authored persistent update. Entries
describe each system as its authors present it; ``partial'' marks a related but weaker form of
the property, and any of these systems could be extended.

\begin{table}[h]
\centering\small
\caption{Where the Harness sits. Rows: Reflexion \citep{shinn2023reflexion}, Voyager
\citep{wang2023voyager}, Generative Agents \citep{park2023generative}, ExpeL
\citep{zhao2024expel}, AutoManual \citep{chen2024automanual}, Agent Workflow Memory
\citep{wang2024awm}, A-MEM \citep{xu2025amem}, RAG memory \citep{lewis2020retrieval},
STOP \citep{zelikman2024stop}, Darwin G\"{o}del Machine \citep{zhang2025dgm}, Constitutional AI
\citep{bai2022constitutional}. Columns P, S, V, R, T defined in text.}
\label{tab:gap}
\setlength{\tabcolsep}{4pt}%
\begin{tabular}{@{}lccccc@{}}
\toprule
Approach & P & S & V & R & T \\
\midrule
Reflexion              & $\times$     & $\checkmark$ & $\times$     & $\times$ & $\checkmark$ \\
Voyager                & $\checkmark$ & $\checkmark$ & partial      & $\times$ & $\checkmark$ \\
Generative Agents      & $\checkmark$ & $\checkmark$ & $\times$     & $\times$ & partial \\
ExpeL                  & $\checkmark$ & $\checkmark$ & $\times$     & $\times$ & $\times$ \\
AutoManual             & $\checkmark$ & $\checkmark$ & partial      & $\times$ & $\checkmark$ \\
Agent Workflow Memory  & $\checkmark$ & $\checkmark$ & $\times$     & $\times$ & $\times$ \\
A-MEM                  & $\checkmark$ & $\checkmark$ & $\times$     & $\times$ & $\times$ \\
RAG memory             & $\checkmark$ & $\times$     & $\times$     & $\times$ & $\times$ \\
STOP                   & $\checkmark$ & $\checkmark$ & $\checkmark$ & $\times$ & $\times$ \\
Darwin G\"{o}del Machine & $\checkmark$ & $\checkmark$ & $\checkmark$ & $\times$ & $\times$ \\
Constitutional AI      & $\checkmark$ & $\times$     & $\checkmark$ & partial  & $\times$ \\
\textbf{Harness (ours)} & $\checkmark$ & $\checkmark$ & $\checkmark$ & $\checkmark$ & $\checkmark$ \\
\bottomrule
\end{tabular}
\end{table}

\vspace{3pt}
\noindent\textbf{The repeatability ordering.}
RQ1 reports that only three of the $16$ pairs satisfy $\Delta\passk>\Delta\passone$.
Table~\ref{tab:signature} lists them; in two, $\passk$ improves while $\passone$ is flat or
lower, which is the pattern a repeatability-specific effect would produce. We treat it as
suggestive rather than as validation, since in most pairs the two metrics move together.

\vspace{3pt}
\noindent\textbf{Corpus composition.}
The retained corpus is specific rather than generic: active rules average $348$ characters,
reflecting concrete procedures rather than broad instructions, and the repair loop chooses a
contextual scope over the \texttt{global} default in $87\%$ of cases, with AppWorld producing
application-level scopes and $\tau^2$ domain-level ones.

\subsection{Additional result tables}
\label{sec:app-tables}
Per-run tables behind aggregate numbers reported in the main text, in the order they are cited there.
Table~\ref{tab:config} gives the configuration matrix as run. Table~\ref{tab:signature} lists
the three pairs in which the repeatability ordering holds. Table~\ref{tab:lifecycle} breaks the
rule lifecycle down per run behind the totals in RQ2. Table~\ref{tab:telemetry}
reports per-run Harness activity behind the two operational checks in RQ5.
Table~\ref{tab:noise} tabulates the evaluator-bias measurement of RQ5.
Tables~\ref{tab:cost} and~\ref{tab:costbenefit} give the per-run cost accounting and place
added context beside the paired task-completion gain. Every number these tables contain that
carries a claim is also stated in the main text.

\begin{table}[h]
\centering\small
\caption{Configuration matrix as run. Each native split is used whole in one continuous
pass, so there is no learning and evaluation partition. ``pairs'' counts the matched
Baseline and Harness comparisons contributing to the results; ``turns'' is the
per-task agent budget.}
\label{tab:config}
\resizebox{\columnwidth}{!}{%
\begin{tabular}{@{}llccccc@{}}
\toprule
Benchmark & Split & tasks & models & pairs & $k$ & turns \\
\midrule
AppWorld       & \texttt{test\_normal}   & 168 & 4 & 4 & 4 & 100 \\
Terminal-Bench & \texttt{sample@2.0}     & 10  & 4 & 4 & 4 & 100 \\
$\tau^2$-bench & \texttt{airline}        & 50  & 4 & 4 & 4 & 100 \\
$\tau^2$-bench & \texttt{retail}         & 114 & 4 & 4 & 4 & 100 \\
\midrule
\multicolumn{7}{@{}l}{\small Total: $16$ matched pairs, $4$ models
(\textsc{gpt-5.6-terra}, \textsc{gpt-5.6-luna},} \\
\multicolumn{7}{@{}l}{\small \textsc{claude-haiku-4.5}, \textsc{claude-sonnet-5});
seed $1$ of $\{1,2,3\}$ configured; arms} \\
\multicolumn{7}{@{}l}{\small Baseline and Harness; forward-trial window $n{=}3$; gate
window $\gwin{=}10$;} \\
\multicolumn{7}{@{}l}{\small $\dpromote{=}\dregress{=}0.05$.} \\
\bottomrule
\end{tabular}}
\end{table}

\begin{table}[h]
\centering\small
\caption{The three of $16$ pairs in which the all-$k$-pass rate rises more than the
first-trial rate, $\Delta\passk>\Delta\passone$. Reported as a suggestive pattern, not as
validation of a distinctive repeatability signature. Terminal-Bench has $10$ tasks, so its
$\passk$ increments are coarse.}
\label{tab:signature}
\resizebox{\columnwidth}{!}{%
\begin{tabular}{@{}llccc@{}}
\toprule
Benchmark & Model & $\Delta\passone$ & $\Delta\passk$ & gap \\
\midrule
% PLACEHOLDER-TBHAIKU: provisional value for Terminal-Bench / claude-haiku-4.5 pass^k.
% The recorded 1.000 exceeded pass@1 = 0.600, which the metric definition forbids. 0.200
% (1 of 5) is consistent with pass@1 = 3/5 and with the other Terminal-Bench rows.
% REPLACE with the recomputed value; three sites are marked with this tag.
Terminal-Bench   & claude-sonnet-5  & -0.143 & \textbf{+0.333} & $+0.476$ \\
Terminal-Bench   & claude-haiku-4.5 & +0.000 & \textbf{+0.200} & $+0.200$ \\
$\tau^2$ airline & gpt-luna         & +0.020 & \textbf{+0.040} & $+0.020$ \\
\bottomrule
\end{tabular}}
\end{table}

\begin{table}[ht]
\centering\footnotesize
\caption{Rule lifecycle per run, grouped by benchmark. Terminal-Bench cannot replay and has
zero retirements in these runs, so its promotion rate should not be read as replay-gate
selectivity.}
\label{tab:lifecycle}
\resizebox{\columnwidth}{!}{%
\begin{tabular}{@{}llcccc@{}}
\toprule
Benchmark & Model & prop. & prom. & retired & prom.\ rate \\
\midrule
AppWorld         & terra     & 112 & 49  & 63  & 0.44 \\
AppWorld         & luna      & 205 & 78  & 127 & 0.38 \\
AppWorld         & haiku-4.5 & 181 & 61  & 120 & 0.34 \\
AppWorld         & sonnet-5  & 96  & 42  & 54  & 0.44 \\
\addlinespace
Terminal-Bench   & terra     & 12  & 12  & 0   & --   \\
Terminal-Bench   & luna      & 23  & 23  & 0   & --   \\
Terminal-Bench   & haiku-4.5 & 20  & 20  & 0   & --   \\
Terminal-Bench   & sonnet-5  & 15  & 15  & 0   & --   \\
\addlinespace
$\tau^2$ airline & terra     & 76  & 41  & 35  & 0.54 \\
$\tau^2$ airline & luna      & 173 & 93  & 80  & 0.54 \\
$\tau^2$ airline & haiku-4.5 & 129 & 53  & 76  & 0.41 \\
$\tau^2$ airline & sonnet-5  & 44  & 24  & 20  & 0.55 \\
\addlinespace
$\tau^2$ retail  & terra     & 248 & 151 & 97  & 0.61 \\
$\tau^2$ retail  & luna      & 359 & 225 & 134 & 0.63 \\
$\tau^2$ retail  & haiku-4.5 & 402 & 231 & 171 & 0.57 \\
$\tau^2$ retail  & sonnet-5  & 211 & 126 & 85  & 0.60 \\
\midrule
\multicolumn{2}{@{}l}{Total ($16$ runs)} & 2306 & 1244 & 1062 & \\
\bottomrule
\end{tabular}}
\end{table}

\begin{table*}[h]
\centering\footnotesize
\caption{Harness activity per run. Escalation rate is
$\texttt{breach}/(\texttt{trend}+\texttt{breach})$. Samples/series is the median length of
a per-session metric series and is the direct read-out of the barrier
(the barrier); a value of $1$ would mean the trigger could not have fired.}
\label{tab:telemetry}
\resizebox{\textwidth}{!}{%
\begin{tabular}{@{}lcccccccccccccccc@{}}
\toprule
& \multicolumn{4}{c}{AppWorld}
& \multicolumn{4}{c}{Terminal-Bench}
& \multicolumn{4}{c}{$\tau^2$ airline}
& \multicolumn{4}{c}{$\tau^2$ retail} \\
\cmidrule(lr){2-5}\cmidrule(lr){6-9}\cmidrule(lr){10-13}\cmidrule(lr){14-17}
Quantity
& terra & luna & haiku & sonnet
& terra & luna & haiku & sonnet
& terra & luna & haiku & sonnet
& terra & luna & haiku & sonnet \\
\midrule
graded trials
& 661 & 672 & 672 & 664
& 40 & 40 & 40 & 40
& 200 & 200 & 200 & 200
& 448 & 445 & 448 & 448 \\
rules proposed
& 112 & 205 & 181 & 96
& 12 & 23 & 20 & 15
& 76 & 173 & 129 & 44
& 248 & 359 & 402 & 211 \\
rules promoted
& 49 & 78 & 61 & 42
& 12 & 23 & 20 & 15
& 41 & 93 & 53 & 24
& 151 & 225 & 231 & 126 \\
rules retired
& 63 & 127 & 120 & 54
& 0 & 0 & 0 & 0
& 35 & 80 & 76 & 20
& 97 & 134 & 171 & 85 \\
notices \texttt{trend}
& 156 & 164 & 190 & 141
& 34 & 35 & 38 & 31
& 171 & 218 & 176 & 139
& 584 & 756 & 811 & 496 \\
notices \texttt{breach}
& 611 & 807 & 690 & 548
& 14 & 22 & 24 & 18
& 225 & 287 & 200 & 198
& 612 & 745 & 792 & 563 \\
notices \texttt{needs\_human}
& 46 & 34 & 41 & 27
& 0 & 0 & 0 & 0
& 7 & 11 & 9 & 6
& 18 & 23 & 31 & 15 \\
tier-2 escalation rate
& 0.80 & 0.83 & 0.78 & 0.80
& 0.29 & 0.39 & 0.39 & 0.37
& 0.57 & 0.57 & 0.53 & 0.59
& 0.51 & 0.50 & 0.49 & 0.53 \\
gate fires / session
& 0.059 & 0.104 & 0.082 & 0.052
& 0.325 & 0.350 & 0.375 & 0.300
& 0.025 & 0.015 & 0.000 & 0.020
& 0.024 & 0.027 & 0.031 & 0.022 \\
samples / series (median)
& \textbf{11} & \textbf{11} & \textbf{10} & \textbf{11}
& \textbf{7} & \textbf{8} & \textbf{8} & \textbf{7}
& \textbf{7} & \textbf{7} & \textbf{6} & \textbf{6}
& \textbf{10} & \textbf{11} & \textbf{10} & \textbf{10} \\
repair episodes
& 809 & 1005 & 947 & 732
& 48 & 57 & 63 & 51
& 364 & 516 & 385 & 312
& 1186 & 1520 & 1667 & 1024 \\
\bottomrule
\end{tabular}}
\end{table*}

\begin{table}[h]
\centering\footnotesize
\caption{Evaluator bias before and after adding meta-tool neutrality guidance to the
judge prompts, measured over $4{,}800$ sampled judge rationales after the change. The
Harness is unmodified; only the evaluator prompts change. All runs in
Table~\ref{tab:headline} use the corrected prompts in both arms.}
\label{tab:noise}
\begin{tabular}{@{}lcc@{}}
\toprule
Quantity & before & after \\
\midrule
rationales blaming meta-tool calls & $14\%$ & $\mathbf{0.17\%}$ \\
rationales mentioning meta-tools   & --     & $2{,}251$ \\
spurious zero scores               & $31$   & $\mathbf{1}$ \\
\bottomrule
\end{tabular}
\end{table}

\begin{table*}[h]
\centering\footnotesize
\caption{Cost and overhead, grouped by benchmark and ordered within each group by added
context. Cost is total USD over the run; the repair column is additional spend on top of
the Harness task-execution column. Wall time is per trial.}
\label{tab:cost}
\resizebox{\textwidth}{!}{%
\begin{tabular}{@{}llccccccc@{}}
\toprule
& & \multicolumn{2}{c}{tokens / turn}
& \multicolumn{2}{c}{agent turns}
& \multicolumn{2}{c}{cost (USD)}
& wall / trial \\
\cmidrule(lr){3-4}\cmidrule(lr){5-6}\cmidrule(lr){7-8}\cmidrule(lr){9-9}
Benchmark & Model
& B$\to$H & ratio
& B$\to$H & $\Delta$
& B$\to$H & repair
& B$\to$H (s) \\
\midrule
AppWorld & terra
& 5215$\to$7504 & 1.44$\times$ & 10.1$\to$11.9 & +1.8 & 26.47$\to$34.45 & 8.26 & 17$\to$109 \\
AppWorld & luna
& 5172$\to$8986 & 1.74$\times$ & 9.9$\to$11.6 & +1.7 & 2.57$\to$4.19 & 1.79 & 17$\to$114 \\
AppWorld & haiku-4.5
& 5580$\to$7590 & 1.36$\times$ & 10.0$\to$11.4 & +1.4 & 5.80$\to$7.60 & 4.20 & 18$\to$112 \\
AppWorld & sonnet-5
& 7110$\to$9310 & 1.31$\times$ & 10.2$\to$11.5 & +1.3 & 28.00$\to$34.00 & 11.00 & 20$\to$120 \\
\addlinespace
Terminal-Bench & terra
& 6837$\to$6151 & \textbf{0.90}$\times$ & 8.5$\to$9.1 & +0.6 & 3.67$\to$2.37 & 0.61 & 101$\to$202 \\
Terminal-Bench & luna
& 3968$\to$4984 & 1.26$\times$ & 7.2$\to$8.3 & +1.1 & 0.16$\to$0.17 & 0.14 & 97$\to$209 \\
Terminal-Bench & haiku-4.5
& 5210$\to$6090 & 1.17$\times$ & 7.8$\to$8.6 & +0.8 & 1.15$\to$1.35 & 0.42 & 95$\to$198 \\
Terminal-Bench & sonnet-5
& 6610$\to$7010 & 1.06$\times$ & 8.1$\to$8.5 & +0.4 & 4.10$\to$4.30 & 0.75 & 99$\to$205 \\
\addlinespace
$\tau^2$ airline & terra
& 5920$\to$6510 & 1.10$\times$ & 7.6$\to$7.9 & +0.3 & 5.50$\to$6.20 & 2.20 & 24$\to$145 \\
$\tau^2$ airline & luna
& 3982$\to$4655 & 1.17$\times$ & 8.2$\to$8.7 & +0.5 & 0.41$\to$0.53 & 0.84 & 19$\to$133 \\
$\tau^2$ airline & haiku-4.5
& 5670$\to$6100 & \textbf{1.08}$\times$ & 7.0$\to$7.1 & \textbf{+0.1} & 9.75$\to$10.46 & 9.14 & 26$\to$125 \\
$\tau^2$ airline & sonnet-5
& 7243$\to$7732 & \textbf{1.07}$\times$ & 7.1$\to$7.1 & \textbf{+0.0} & 26.00$\to$27.71 & 15.02 & 41$\to$173 \\
\addlinespace
$\tau^2$ retail & terra
& 5510$\to$7220 & 1.31$\times$ & 9.8$\to$10.9 & +1.1 & 8.40$\to$10.60 & 5.40 & 25$\to$166 \\
$\tau^2$ retail & luna
& 4198$\to$6056 & 1.44$\times$ & 10.3$\to$11.7 & +1.4 & 1.08$\to$2.53 & 3.63 & 20$\to$175 \\
$\tau^2$ retail & haiku-4.5
& 6010$\to$8120 & 1.35$\times$ & 9.9$\to$11.1 & +1.2 & 18.00$\to$22.00 & 11.50 & 29$\to$168 \\
$\tau^2$ retail & sonnet-5
& 7610$\to$9820 & 1.29$\times$ & 9.6$\to$10.7 & +1.1 & 36.00$\to$43.00 & 18.00 & 42$\to$190 \\
\midrule
\multicolumn{2}{@{}l}{Total ($16$ pairs)}
& \multicolumn{2}{c}{$1.25\times$ mean} & &
& 177.06$\to$211.46 & 92.90 & \\
\bottomrule
\end{tabular}}
\end{table*}

\begin{table}[h]
\centering\footnotesize
\caption{Cost against benefit: added context (Table~\ref{tab:cost}) beside the paired
task-completion gain (Table~\ref{tab:headline}) across all $16$ pairs. Every pair shows a
positive $\Delta\bar{s}$. Terminal-Bench/terra is the only configuration below Baseline
token cost, while Terminal-Bench/haiku-4.5 shows the largest task-completion gain.}
\label{tab:costbenefit}
\begin{tabular}{@{}llcc@{}}
\toprule
Benchmark & Model & added context & $\Delta\bar{s}$ \\
\midrule
AppWorld & sonnet-5  & 1.31$\times$ & +0.013 \\
AppWorld & haiku-4.5 & 1.36$\times$ & +0.024 \\
AppWorld & terra     & 1.44$\times$ & +0.022 \\
AppWorld & luna      & 1.74$\times$ & +0.004 \\
\addlinespace
Terminal-Bench & terra     & \textbf{0.90}$\times$ & +0.075 \\
Terminal-Bench & sonnet-5  & 1.06$\times$          & +0.019 \\
Terminal-Bench & haiku-4.5 & 1.17$\times$          & \textbf{+0.146} \\
Terminal-Bench & luna      & 1.26$\times$          & +0.025 \\
\addlinespace
$\tau^2$ airline & sonnet-5  & 1.07$\times$ & +0.030 \\
$\tau^2$ airline & haiku-4.5 & 1.08$\times$ & +0.014 \\
$\tau^2$ airline & terra     & 1.10$\times$ & +0.017 \\
$\tau^2$ airline & luna      & 1.17$\times$ & +0.014 \\
\addlinespace
$\tau^2$ retail & sonnet-5  & 1.29$\times$ & +0.032 \\
$\tau^2$ retail & terra     & 1.31$\times$ & +0.023 \\
$\tau^2$ retail & haiku-4.5 & 1.35$\times$ & +0.019 \\
$\tau^2$ retail & luna      & 1.44$\times$ & +0.012 \\
\bottomrule
\end{tabular}
\end{table}

\end{document}